***AI Literacy for Legal Translation: Developing Digital Resilience***

Łucja Biel
l.biel@uw.edu.pl
University of Warsaw

**Abstract:** Generative AI is transforming legal translation by introducing opportunities alongside linguistic, technical, legal, ethical and cognitive risks. This chapter examines the implications of AI for professional legal translation and proposes an AI literacy framework tailored to the profession. It argues that AI does not change the fundamental objectives of legal translation but requires an extension of professional competence through AI literacy. The proposed framework comprises four mutually reinforcing dimensions — foundational, procedural, critical and strategic — and conceptualises AI literacy as a transversal component of legal translation competence that fosters digital resilience. It further discusses the pedagogical implications of this framework by proposing classroom activities designed to develop AI literacy in legal translator education, enabling future translators to integrate AI critically, responsibly and in accordance with professional standards.

The translation industry is undergoing one of the most profound and disruptive transformations in its history. According to the *2026 European Language Industry Survey Report*, translators are facing mounting professional pressures and economic precarity. Driven by the widespread uptake of artificial intelligence (AI) and the accompanying hype, traditional human translation is being replaced by workflows with machine translation and generative artificial intelligence (GenAI). As clients turn to raw machine translation for routine content, demand for conventional translation services is declining. Where human intervention remains necessary, clients tend to demand post-editing to secure faster turnaround times and lower costs, while continuing to expect quality comparable to that of human translation (ISO 2017). These developments have eroded confidence across the profession and raised concerns about its long-term sustainability (ELIS 2026).

These market pressures are perhaps felt less acutely in legal translation but they certainly exist. Legal translation has traditionally been among the most automation-resistant sectors of the translation industry. Concerns about hallucinations, the opacity of AI systems, confidentiality, accountability and professional trust have contributed to a slower adoption of AI than in other domains (Dahl et al. (2024), Biel, Scott, and O'Shea (2024), Girard, Lehoux-Jobin, and Pomerleau (2026)). Nevertheless, AI is reshaping expectations of legal translators' professional practice. Legal translation has gradually shifted towards hybrid workflows as legal translators are expected to work with machine translation (Killman (2024), Prieto Ramos (2025)), and, more recently, GenAI. Yet the adoption of AI in legal translation cannot be assessed solely in terms of efficiency gains. Legal translation operates in high-risk settings where translation errors may have severe consequences. The growing use of AI therefore reinforces, rather than diminishes, the importance of informed human judgement.

This chapter argues that these developments make AI literacy an essential transversal layer of legal translation competence. It proposes a framework of AI literacy designed for legal translation, which takes into account its AI-related risk profile. The chapter concludes by outlining practical approaches to developing AI literacy in legal translator education, supported by examples of classroom activities.

## 1. Legal translation: Professional responsibility and trust

Law is fundamentally a form of communication, serving as a system of binding rules that regulates society and structures interactions among governments, organisations, legal professionals, businesses, and individuals (van Hoecke 2002, 7). In multilingual contexts, this communication depends on legal translation, which enables communication across linguistic and jurisdictional boundaries, serving normative, informational, judicial and general legal purposes (Cao 2007). In doing so, it performs essential societal, economic and political functions. It contributes to access to justice, fairness, and social inclusion by ensuring that legal rights and obligations can be exercised irrespective of language. At the same time, it facilitates international trade and transnational governance in an increasingly interconnected world (Biel and Laforcade 2026, 112).

The significance of this responsibility becomes apparent where legal translation is not only a professional service but also a legal entitlement and a safeguard of fundamental rights. This can be illustrated by the protection of linguistic minorities and of persons involved in court proceedings (Doczekalska and Biel 2022). Language constitutes one of the distinctive features of ethno-cultural minorities (Wheatley 2005), and the Universal Declaration of Human Rights (1948) prohibits discrimination on the grounds of language. In some jurisdictions, this entails the right to communicate with public authorities in minority languages — a protection that depends on the provision of translation and interpreting. Translation and interpreting are also recognized as procedural guarantees that enable suspected and accused persons to exercise their right to a fair trial. They, for example, guarantee that such individuals receive translation of critical documents, e.g. charges, detentions, and judgments. This principle has been enshrined in various international frameworks, including the European Convention on Human Rights and Fundamental Freedoms of 1950[1], the International Covenant on Civil and Political Rights of 1966[2], and Directive 2010/64/EU of the European Parliament and the Council of 20 October 2010 on the right to interpretation and translation in criminal proceedings[3].

The responsibility attached to these functions is further heightened by the inherent complexity of legal translation. Legal translators mediate not only between languages but also between legal systems, legal cultures and institutional practices. Despite plain language initiatives (see e.g. Williams (2022), ISO (2025)), legal language remains conceptually and syntactically complex. The principal challenge lies in the incongruity of terminology across legal systems because it derives its meaning from specific jurisdictions (Šarčević (1997), Biel (2022)). Furthermore, the open-textured nature of law (Hart 1994) introduces deliberate flexibility and ambiguities designed to accommodate judicial interpretation and future events. At the same time, the translation of legal texts demands a high degree of accuracy, as even minor shifts in meaning can alter legal rights, obligations, or procedural outcomes, potentially resulting in litigation, contractual disputes, financial losses, reputational damage, or violation of rights (Scott and O'Shea 2021). This tension between accuracy and flexibility requires expert human judgment.

Given this interplay between the complexity, functions and consequences, legal translation depends on high levels of professional competence and public trust. It frequently operates within institutional settings where translators may work as civil servants and/or perform functions associated with professions of public trust, including sworn or certified translators. This status is reflected in international standardisation: the legal translation standard ISO 20771 (ISO 2020) specifies much higher requirements for legal translators than the general translation standard ISO 17100. These include formal qualifications in translation, law or related disciplines, substantial professional experience in legal translation, postgraduate qualifications and continuing professional development (ISO 2020). In many jurisdictions, legal translators also bear professional and, in some cases, legal responsibility for the quality of their work through certification schemes, statutory regulation or professional codes of ethics.

[1] https://www.echr.coe.int/documents/d/echr/convention_ENG
[2] https://www.ohchr.org/en/instruments-mechanisms/instruments/international-covenant-civil-and-political-rights
[3] https://eur-lex.europa.eu/eli/dir/2010/64/oj/eng

## 2. AI in legal translation: Between augmentation and automation

This section explores how AI has been transforming legal translation practice, arguing that legal accountability requires human-centered augmentation rather than unsupervised automation. It first details the evolution toward hybrid workflows and the cautious AI adoption among practitioners. It then maps the AI-related risk profile of legal translation, demonstrating why commercial "good enough" quality metrics are problematic in legal communication.

### 2.1 The augmented legal translator

Technological support is not new to legal translation. Long before the emergence of GenAI, working in technologically mediated environments was standard practice for legal translators, who integrated digital resources into their daily work. These include computer-assisted translation (CAT) tools, translation memories, terminology management systems, quality assurance and revision software, corpora, databases of legal texts (e.g. Eur-Lex[4], Curia[5]) and multilingual termbases, such as the European Union's IATE[6] or the United Nation's UNTERM[7]. Such technologies facilitate information mining, terminology management, consistency checks and the reuse of translations, and, hence, support both translation quality and productivity. They have traditionally functioned as administrative support, assisting research and professional decision-making while leaving translation choices with the translator. In O'Brien's (2024, 5) terms, these technologies exemplify augmentation: they do not substitute for but support human problem solving.

The current generation of AI systems represents a different stage in this technological evolution. Neural machine translation (NMT) engines, such as generalist DeepL, Google Translate or customised institutional eTranslation[8], have become an established component of many translation workflows. More recently, large language models (LLMs), e.g. ChatGPT, Claude and Gemini, have introduced new possibilities for generating, translating and revising text, as well as for auxiliary tasks (Bogucki 2026). Unlike earlier technologies, these systems can support a much wider range of translation-related tasks, including activities that previously relied almost entirely on human expertise. Consequently, legal translation has increasingly adopted hybrid workflows that combine translation memories, machine translation, LLMs and human post-editing within CAT environments. While these developments have improved productivity, they have also intensified commercial pressures to reduce turnaround times and costs, reinforcing the market-wide transition towards AI-assisted production (Martinez Carrasco, Borja Albi, and Biel 2024).

This shift reflects a broader transformation extending beyond translation into legal practice itself. AI is becoming embedded across the legal sector through the rapid expansion of

[4] https://eur-lex.europa.eu/
[5] https://infocuria.curia.europa.eu/
[6] https://iate.europa.eu/
[7] https://unterm.un.org/unterm2/
[8] https://language-tools.ec.europa.eu

LegalTech[9]. Recent years have witnessed an unprecedented proliferation of GenAI-powered legal technologies supporting legal research, contract drafting, document review, summarizing, litigation management, compliance, legal analytics and organisational workflows (LegalTech Hub 2026). As routine legal tasks are more and more automated, multilingual communication also becomes subject to growing expectations of automation. The consequence is that legal translators are no longer responding solely to technological developments within the language industry but also to changing expectations originating in the legal profession.

This development necessitates a distinction between augmentation and automation. Augmentation refers to the use of AI to support professional judgement while preserving human control over translation decisions; it amplifies human intellectual abilities, leading to human-centered augmented translation (O'Brien 2024). Automation, by contrast, seeks to replace human decision-making by relying on unsupervised AI. The distinction is not merely technological but conceptual. Whereas augmentation assumes that AI assists the translator, automation assumes that AI can replace the translator. As argued in the previous section, legal translation is characterised by professional responsibility and legal accountability. Whether automation can satisfy these requirements therefore becomes a central question.

## 2.2 AI adoption in legal translation: Cautious augmentation

While legal translators widely use translation technologies, they seem to integrate automated systems cautiously and selectively. Surveys report a widespread use of translation memories and growing integration of machine translation within institutional environments (e.g. Cadwell et al. (2016), Rossi and Chevrot (2019), Lesznyák (2019), Prieto Ramos (2024a)), where customised systems and technology-oriented training are readily available (Svoboda, Biel, and Sosoni 2023). For example, a 2023 survey of 474 respondents from 24 institutions (Prieto Ramos 2024a) reports an NMT uptake at the level of ca. 80% for legal translation. A much lower MT uptake at the level of 41% was reported among UK-based freelance translators (Góngora-Goloubintseff 2026)[10]. Nevertheless, machine translation is (still) generally treated as a source of translation suggestions rather than an autonomous producer of legal texts as claimed by ill-informed media narratives[11]. Institutional translators rely primarily on translation memories when verifying previous solutions and post-edit machine translation selectively, in particular when translating legally binding documents (Prieto Ramos 2025). The slightly lower uptake of machine translation for legal texts than for institutional translation more generally (Prieto Ramos 2025) further illustrates translators' awareness of the risks associated with legal communication.

While empirical studies of GenAI adoption among legal translators are not yet available, general industry data suggests that GenAI integration remains limited for core translation tasks and is largely restricted to supporting roles. Translators distinguish between NMT, which is viewed as a specialised translation technology, and GenAI / LLMs, which are primarily deployed as auxiliary aids (Rivas Ginel and Moorkens 2025, 289-290)[12]. However, recent findings by Mao and Zheng (2026, 30)[13] indicate higher uptake among practitioners with strong subject-matter expertise and AI literacy, where adoption also extends to translation drafting, though rarely for expertise-intensive tasks. The primary drivers of GenAI integration include productivity gains, enhanced grammatical and terminological precision, stylistic refinement, improved fluency, and inspiration during translation impasses (Mao and Zheng 2026). Accordingly, translators utilise LLMs mainly for terminology consultation, text revision, paraphrasing, and gisting, alongside administrative tasks such as image-to-text conversion, terminology management, and bilingual alignment (Mao and Zheng (2026), Rivas Ginel and Moorkens (2025)). Interestingly, half of translators surveyed by Mao and Zheng (2026) report finding GenAI output easier to revise than NMT. Nevertheless, concerns regarding factual reliability, hallucinations, confidentiality, privacy breaches, intellectual property risks, bias, and environmental costs limit translators' trust in LLM-generated translations (Rivas Ginel and Moorkens (2025), Mao and Zheng (2026)). These concerns may be expected to be amplified in legal translation.

[9] According to Legaltech Hub (2026), the number of GenAI-powered legal technologies reached 1,196 by June 2026, more than doubling in size since 2025.

[10] The survey was conducted on a small sample of translators (n=32) in 2024.

[11] See, for example, Sorgi and Di Sario (2023), who sensationalise a 17% reduction in European Commission translation staff as proof of automation "killing the EU's translators," overlooking broader structural shifts such as outsourcing.

[12] Rivas Ginel and Moorkens' (2025) survey of professional translators of various specialisations (n=252) and translation platform posts captures practices as of 2023; LLM adoption has likely increased since.

[13] Conducted in 2025, the study surveyed 268 translators who use GenAI, 34% of whom identified the legal sector as their primary area of practice (Mao and Zheng 2026).

### 2.3 AI as a source of professional risk: AI-related risk profile of legal translation

The relatively cautious adoption of AI by legal translators reflects the risk profile of legal communication. AI creates, on the one hand, opportunities for productivity gains and, on the other hand, introduces new sources of professional risk. For analytical purposes, I propose an AI-related risk profile of legal translation comprising five interrelated categories according to the aspect of professional practice they affect: linguistic, technical, legal, ethical and cognitive (Figure 1). Although many of these risks were first identified in the context of NMT, they apply equally, and in several respects more acutely, to GenAI.

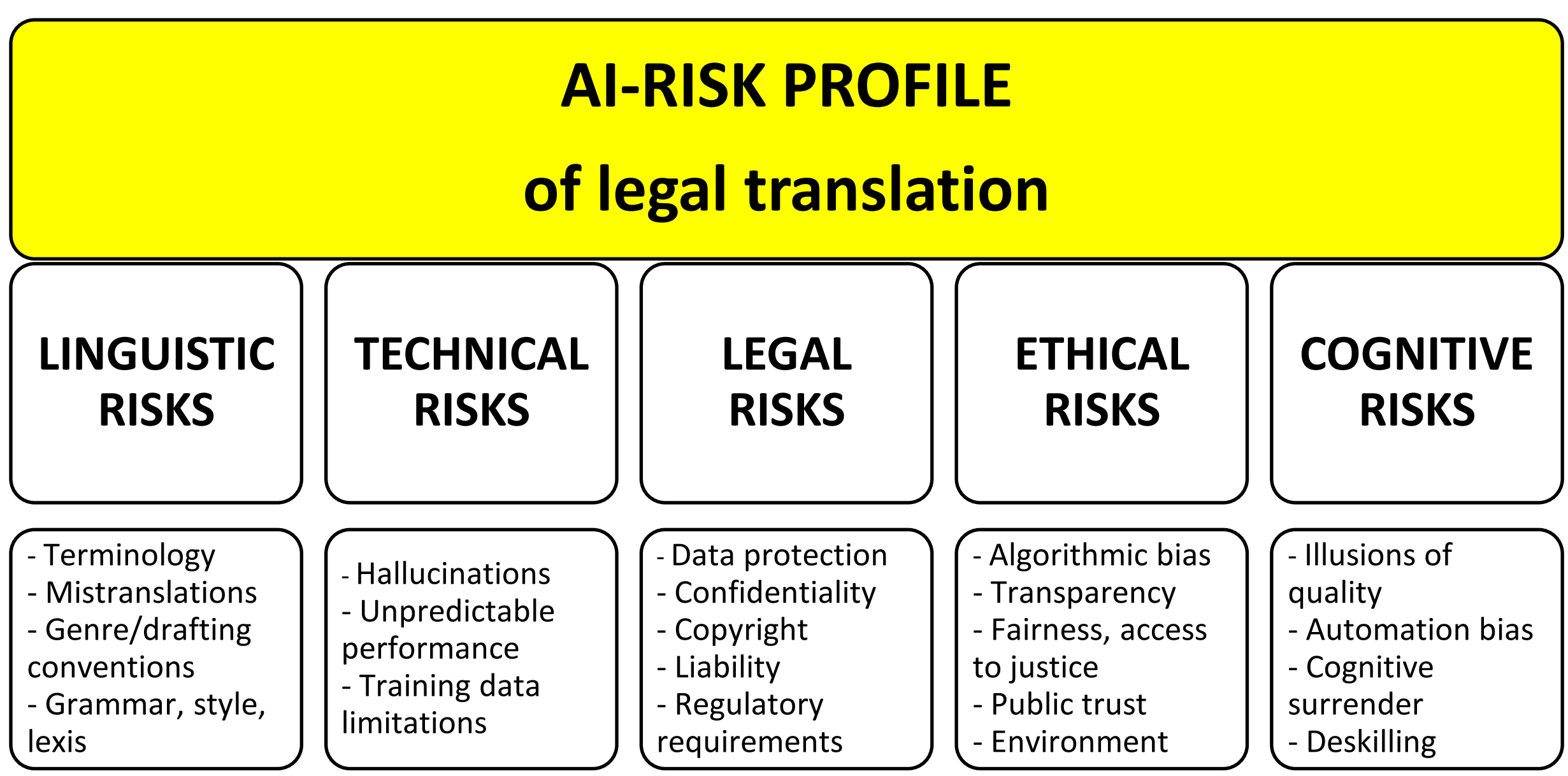

Figure 1: The AI-related risk profile of legal translation: risk categories and their manifestations

The first group comprises **linguistic risks**, which concern the quality of AI-generated translations. The most frequently reported problem is terminological inconsistency and inadequacy, identified by professional translators as one of the main weaknesses of machine translation in legal settings (Lesznyák (2019), Stefaniak (2022), Prieto Ramos (2025), Góngora-Goloubintseff (2026)). Other recurring problems include mistranslations and omissions, non-compliance with institutional drafting and genre conventions, language errors, difficulties in rendering jurisdiction-specific concepts or non-translatables (cf. Lesznyák (2019), Stefaniak (2020), Biel (2021), Prieto Ramos (2025)). While contemporary NMT systems have substantially improved translation quality and increasingly support controlled terminology[14], LLMs introduce additional challenges. Their probabilistic text generation may produce inconsistent terminology, over-fluent but legally inaccurate formulations, or fabricated information (hallucinations), making errors more difficult to detect despite greater linguistic naturalness (Bajčić and Golenko 2024)[15].

The second group concerns **technical risks**, which originate in the way AI systems are developed and trained. Both NMT and LLMs are probabilistic models whose performance depends on the quality, quantity and representativeness of training data. Thus, translation quality varies across languages, domains and text types, with lower-resourced legal systems remaining particularly challenging (Bajčić and Golenko (2024), Góngora-Goloubintseff (2026)). This variability makes AI performance difficult to predict. Stefaniak's (2022) large-scale evaluation of the European Commission's eTranslation across the EU's 24 official languages demonstrated high variation in the treatment of legal terminology depending on language pair, legal genre and term characteristics. Another widely discussed example of GenAI technical limitations is hallucination. Hallucinations have been observed in NMT but are more pronounced in LLMs, sometimes referred to as stochastic parrots, as they are pretrained to generate confident, statistically probable responses rather than admit uncertainty (Kalai et al. 2025). In the legal domain, Dahl et al. (2024) found that leading LLMs hallucinated legal authorities in 58%-88% of tested cases while failing to recognise incorrect legal assumptions in users' prompts.

The third category comprises **legal risks**[16], which concern compliance with legal obligations and regulatory requirements and, hence, the legal permissibility of AI use. Legal translation frequently involves confidential information protected by professional secrecy, contractual obligations and data protection legislation. Uploading privileged documents to publicly accessible or opaque AI systems may expose translators and their clients to breaches of confidentiality, non-disclosure agreements and data protection legislation[17]. Additional legal uncertainty arises from ongoing disputes concerning the use of copyrighted material for AI training, the ownership of AI-generated content, and emerging liability regimes governing AI-assisted professional services (Panezi and O'Shea 2025).

The fourth category concerns **ethical risks**, which relate to professional integrity and societal consequences of AI deployment. Legal translation safeguards access to justice, equal treatment before the law and minority protection. Algorithmic bias, limited transparency and the opacity of AI models may adversely affect it by reproducing discriminatory language, privileging dominant legal cultures or distorting legal interpretation. In LLMs, these risks may be heightened by alignment techniques, such as Reinforcement Learning from Human Feedback (RLHF), which may introduce political, cultural, or institutional biases and expose models to manipulation (Moorkens and Doğru 2026, 109). Uncritical reliance on AI may also disproportionately disadvantage vulnerable persons in the justice system (Panezi and O'Shea 2025). Ethical evaluation therefore requires the "triple bottom line" (Moorkens et al. 2024), which balances technological efficiency with broader impacts on society and the environment.

---

[14] For example, eTranslation allows users to upload custom glossaries to enforce domain terminology (https://translation.ec.europa.eu/tools-and-resources/resources-language-professionals_en).

[15] Conversely, Briva-Iglesias, Dogru, and Cavalheiro Camargo (2024) report that LLMs can achieve better terminological consistency and adequacy compared to NMT.

[16] Legal risks, which are typically treated as part of ethical risks, warrant separate consideration in legal translation.

[17] Confidentiality breaches have been reported in both corporate and public sectors: in 2017, Statoil used free online MT, leaving sensitive contracts, dismissal letters, and passwords searchable on Google (NRK 2017), while the use of Google Translate by a Russian intelligence unit allowed US law enforcement to intercept unencrypted server processing logs (Weiss et al. 2026).

Finally, AI introduces **cognitive risks** that affect professional judgement and long-term sustainability of expertise. Contemporary AI systems generate fluent, idiomatic texts that create an "illusion of quality", making accuracy issues more difficult to detect. This apparent fluency may encourage automation bias, whereby users become predisposed to accept AI outputs despite their potential inaccuracies (Rivas Ginel and Moorkens 2025, 286). Over time, repeated reliance on AI may progress from cognitive offloading, i.e. the deliberate delegation of routine processing to technology to free up mental capacity for complex tasks, to cognitive surrender, where critical evaluation is abandoned in favour of passive acceptance of AI outputs. This increases the risk of deskilling (Kim et al. 2026), leading to a stagnation in competence development or even atrophy of core translation competences.

These observations reinforce Pym's (2025) characterisation of translation as a form of risk management. The five categories demonstrate that AI-related risks extend beyond translation quality. They affect the translated text, the technological infrastructure that produces it, the legal and ethical framework within which translators operate, and the cognitive processes governing professional judgement.

### 2.4 Quality: Is "good enough" good enough?

These developments have important implications for the understanding of translation quality. The widespread adoption of NMT and, more recently, the hype surrounding LLMs have gradually shifted the perception of quality as an absolute standard to a functional notion of "fitness for purpose" (ISO 2015). Quality is viewed as a continuum extending from raw machine translation, through varying levels of post-editing, to fully human translation (Biel 2021). In commercial translation markets, this has been accompanied by a growing acceptance of "good enough" quality, reflecting clients' willingness to trade quality for lower costs and faster delivery. The notion of "good enough" is context-dependent and has itself evolved over time. Initially associated with full post-editing, it has expanded to encompass light post-editing and, increasingly, the use of raw machine translation without human intervention, even in high stakes scenarios (Moorkens et al. 2024).

The limitations of the "good enough" paradigm are particularly evident when AI is used by non-translation professionals. Existing research suggests that legal professionals frequently underestimate the limitations of machine translation and are insufficiently aware of its risks (Vieira, O'Hagan, and O'Sullivan 2021). For instance, Setkowicz-Ryszka's (2024) survey of Polish legal practitioners demonstrates a high degree of trust in generic machine translation, alongside a pronounced lack of risk awareness. Similar findings by Kourouni and O'Shea (2025) in Greece and Girard, Lehoux-Jobin, and Pomerleau (2026) in Canada confirm this pattern: legal professionals regularly use freely available NMT with high self-reported satisfaction, yet a substantial proportion perform little to no post-editing. Such findings indicate that user satisfaction is not necessarily correlated with translation quality, especially where users lack translation expertise.

The increasing availability of AI has also encouraged proposals to automate multilingual communication within judicial systems. In Poland, for example, a report issued by the Ministry of Digital Affairs recommends, based on unsubstantiated claims, the use of automatic interpreting in court proceedings and machine translation for selected procedural documents to improve efficiency (GRAI 2023). Similar initiatives have been reported in Greece and the Czech Republic, reflecting what Panezi and O'Shea (2025) describe as "the creeping decentering of the human". These developments demonstrate that GenAI has begun to shape institutional expectations.

Yet judicial practice has demonstrated the limits of such automation, highlighting how uncritical reliance on machine translation may compromise the evidentiary value and admissibility of information in court. In *United States v. Cruz-Zamora*[18] and *United States v. Ramirez-Mendoza*[19], for example, evidence obtained through police interactions mediated by Google Translate was held to be deficient because the application could not guarantee that constitutional consent had been given knowingly and voluntarily. Many legal systems retain requirements for certified human translations precisely because they provide an accountable professional who assumes responsibility for translation. Although attitudes of authorities towards the use of AI remain inconsistent (Panezi and O'Shea (2025), Vieira (2026, 70)), these decisions demonstrate that AI use should also consider evidential reliability, procedural safeguards and accountability.

### 2.5 From automation back to augmentation

The evidence reviewed in this section confirms that AI is becoming an integral part of legal translation practice. The question is therefore no longer whether AI should be used, but how it can be integrated efficiently and responsibly in light of the linguistic, technical, legal, ethical and cognitive risks discussed above. Current evidence supports an augmentation model, in which AI assists legal translators but does not replace their professional judgement or responsibility for translation. Working effectively in such environments requires AI literacy, to which the next section now turns.

## 3. AI literacy and digital resilience in professional legal translation

This section proposes an AI literacy framework tailored to the professional responsibilities of legal translators and argues that AI literacy should be understood as a transversal layer of legal translation competence. By strengthening informed professional judgement, AI literacy provides the foundation for digital resilience.

### 3.1 What are AI literacy and digital resilience?

The growing importance of AI literacy has recently been reinforced by the EU Artificial Intelligence Act of 2024[20], applicable throughout the European Union. Article 4 requires deployers of AI systems to ensure that persons using such systems possess a sufficient level of AI literacy. Translators who use AI systems, including machine translation, for professional purposes, fall within the scope of this obligation. The Act defines AI literacy as the skills, knowledge and understanding that enable informed use of AI systems while taking into account their opportunities, risks and potential harms.

The concept itself predates the AI Act. Long and Magerko's frequently cited definition describes AI literacy as "a set of competencies that enables individuals to critically evaluate AI technologies; communicate and collaborate effectively with AI; and use AI as a tool online, at home, and in the workplace" (2020, 2). Subsequent frameworks have expanded this perspective by emphasising four complementary dimensions (Ng et al. 2021):

(1) know and understand AI;
(2) use and apply AI,

---

[18] *United States v. Cruz-Zamora*, 318 F. Supp. 3d 1264 (D. Kan. 2018).
[19] *United States v. Ramirez-Mendoza*, No. 4:20-CR-00107, 2021 WL 4502266 (M.D. Pa. Oct. 1, 2021).
[20] Regulation (EU) 2024/1689 of the European Parliament and of the Council of 13 June 2024 laying down harmonised rules on artificial intelligence and amending Regulations (EC) No 300/2008, (EU) No 167/2013, (EU) No 168/2013, (EU) 2018/858, (EU) 2018/1139 and (EU) 2019/2144 and Directives 2014/90/EU, (EU) 2016/797 and (EU) 2020/1828, OJ L, 2024/1689, 12.7.2024.

(3) evaluate and create AI ("higher-order thinking skills"), and

(4) AI ethics ("human-centered considerations", such as fairness, accountability, transparency, safety).

Existing conceptualisations converge on a common principle: AI literacy combines technical understanding with critical evaluation and responsible use.

Translation Studies has approached this issue through the concept of digital literacy. With the growing technologisation of the profession, technological competence has long been recognised as a core component of translation competence. The widespread adoption of NMT shifted attention from technical proficiency to critical thinking skills (Bowker 2025). This development gave rise to machine translation literacy, defined for example as "knowing how MT works, how it can be useful in a particular context, and what the implications are of using MT for specific communicative needs" (O'Brien and Ehrensberger-Dow 2020). More recently, Krüger (2024) positioned AI literacy alongside machine translation literacy and data literacy as complementary dimensions of translators' digital literacy (Krüger 2024). In this framework AI literacy is adapted to include technical knowledge, interaction with AI systems, domain-specific performance, implementation and ethical considerations (Krüger 2024).

AI literacy serves a broader purpose: it provides the foundation for digital resilience. Translators working with GenAI report technology-related stress, professional identity erosion and perceptions of skill underutilisation (Mao and Zheng 2026). Digital resilience acts as a psychological buffer, helping translators to cope with digital anxiety created by automation (Kornacki and Pietrzak 2024). Digital resilience can thus be understood as translators' capacity to adapt constructively to rapid technological change while maintaining professional agency, confidence, and well-being (Kornacki and Pietrzak 2024).

### 3.2 AI Literacy Framework for legal translation

Building on the discussion above, I propose an AI literacy framework tailored to professional legal translation. Existing AI literacy frameworks are largely domain-neutral and do not fully account for professional responsibilities associated with legal communication. The framework proposed here adapts general AI literacy frameworks developed by, in particular Ng et al. (2021), but also Long and Magerko (2020) and Krüger (2024), to the requirements of legal translation and its specific risk profile discussed in Section 2.3.

AI literacy for legal translators may be defined as the capacity to make informed, critical and ethically responsible decisions about the use and supervision of AI throughout the legal translation process while preserving translation quality, legal certainty and professional accountability. This definition deliberately shifts the emphasis from operating AI systems towards professional judgement. Accordingly, AI literacy is conceptualised here as a multidimensional construct comprising four mutually reinforcing dimensions: Foundational, Procedural, Critical and Strategic (Figure 2 and Table 1). Each dimension builds on the previous one: from understanding AI systems (Foundational), through integrating them into legal translation workflows (Procedural) and critically evaluating their output (Critical), to exercising professional judgement about whether, when and how AI should be used (Strategic).

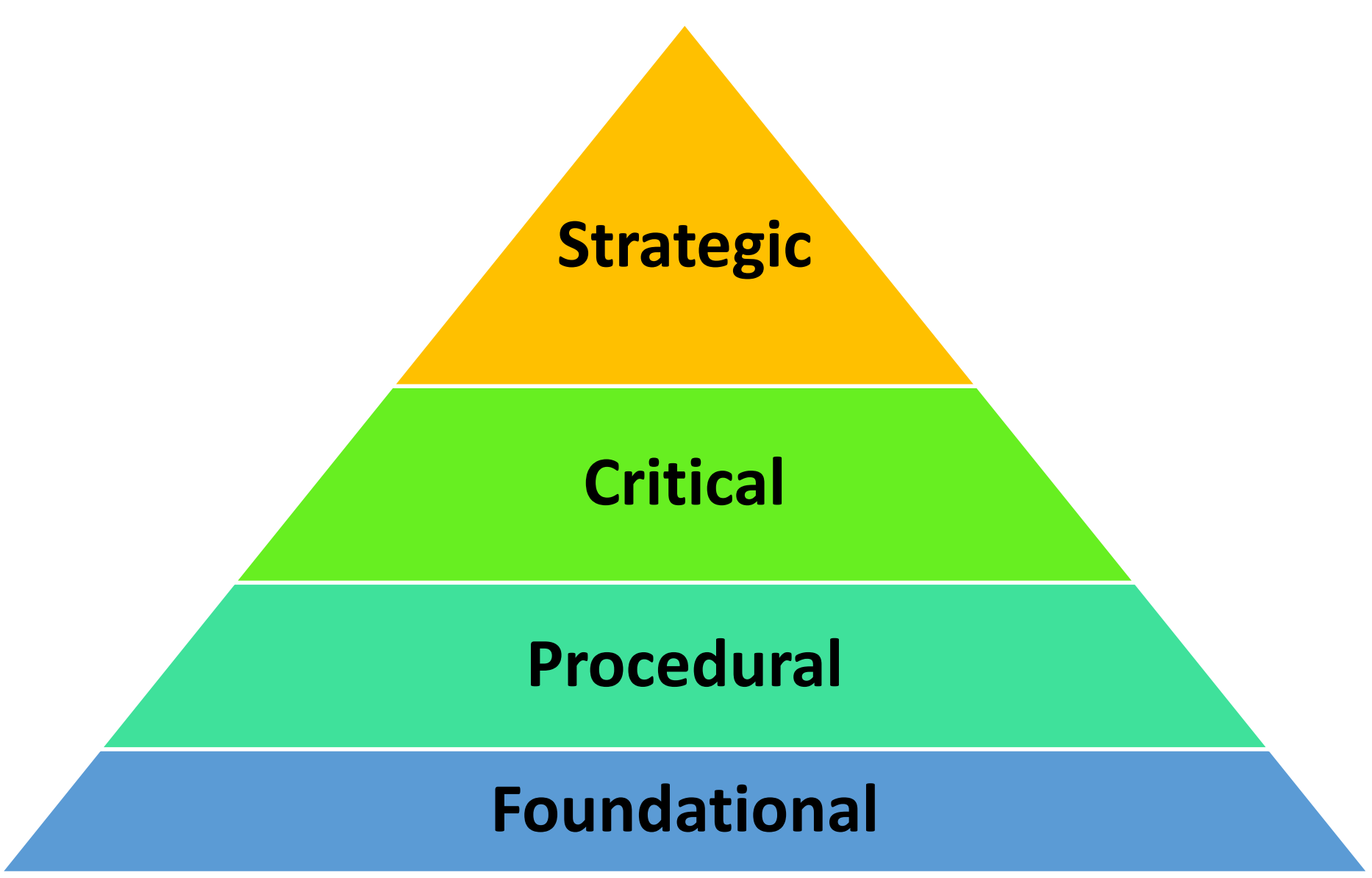

Figure 2: AI Literacy Pyramid for Legal Translation.

The base is the **foundational dimension**, which concerns knowing and understanding the nature and functions of AI systems. Legal translators require a working knowledge of the main technologies used in translation practice, including NMT and LLMs, their probabilistic character, training data, capabilities, limitations and associated risks. This dimension also encompasses data literacy, including awareness of why AI performance differs across languages, legal domains and text types, and why hallucinations and algorithmic bias occur.

Building on this knowledge, the **procedural dimension** concerns the effective integration of AI into translation workflows and interaction with AI systems. It includes selecting appropriate technologies for translation tasks, using AI to support terminology research, information retrieval, drafting and revision, as well as their integration into CAT environments and hybrid workflows. It also covers prompt engineering, that is the capacity to formulate theory-informed and context-aware prompts and conduct iterative interactions with GenAI to refine outputs.

The **critical dimension** introduces the higher-order cognitive skills required for human oversight of AI-assisted translation and quality assurance. It encompasses critical evaluation, post-editing and verification strategies, including the detection of hallucinations, mistranslations, terminological inconsistencies, failures to comply with legal drafting conventions, unnaturalness, as well as other linguistic and legal aspects of translation quality. It also requires awareness of cognitive risks associated with AI-assisted translation, particularly automation bias and illusions of quality.

At the apex of the framework lies the **strategic dimension**, which governs professional judgement. It integrates legal, ethical and reflective dimensions of AI use. It requires translators to assess whether AI should be used for a particular assignment, determine the appropriate degree of human intervention, and justify technology choices in light of the risk profile of legal translation, in particular legal risks (e.g. confidentiality obligations, professional liability), ethical considerations (transparency, fairness, access to justice), as well as client expectations. It also encompasses reflective practice aimed at maintaining professional competence, thereby supporting professional agency and digital resilience.

Table 1. AI Literacy Framework for legal translation: dimension, functions and components

| DIMENSION | FUNCTION | KNOWLEDGE AND SKILLS |
| --- | --- | --- |
| **Strategic** *(Why? When?)* | Professional judgement and governance of AI use | Risk assessment; technology selection; workflow planning; legal and ethical compliance; client communication; justification of AI use; reflective practice; professional agency |
| **Critical** *(Can I trust this?)* | Human oversight and quality assurance | Verification strategies; post-editing; translation quality assessment; hallucination detection; identification of errors; automation bias awareness |
| **Procedural** *(How?)* | Human-AI interaction and workflow integration | Prompt engineering; AI-assisted terminology research and information retrieval; AI-supported drafting and revision; CAT-NMT-LLM integration; workflow management |
| **Foundational** *(What?)* | Understanding AI | AI concepts; NMT and LLMs; probabilistic generation; training data; capabilities and limitations; data literacy |

The four dimensions of AI literacy are mutually reinforcing. Foundational knowledge enables effective interaction with AI systems; procedural competence creates the conditions for human oversight; critical evaluation informs strategic decision-making, while strategic judgement continuously shapes procedural choices throughout the translation process. Together, these dimensions enable legal translators to integrate AI into professional practice responsibly, critically and ethically. AI literacy also provides the basis for digital resilience. By supporting informed technology adoption and critical reflection, it enables legal translators to adapt to technological change while maintaining professional autonomy and confidence.

### 3.3 AI literacy as a transversal layer of legal translation competence

The growing integration of AI into legal translation workflows calls for updating existing legal translation competence models, most of which predate the widespread adoption of NMT and GenAI. Although AI-assisted workflows change the conditions under which translators work, they do not alter the central premise of these models: high-quality legal translation depends on the coordinated deployment of strategic, thematic, linguistic, instrumental and professional competences. Indeed, AI reinforces their importance because they enable translators to evaluate AI-generated output and assume responsibility for the final translation. The AI literacy framework proposed in section 3.2 complements these competences by equipping translators to exercise them effectively in AI-assisted environments.

Existing competence models, including those proposed by Scarpa and Orlando (2017), ISO (2020), Prieto Ramos (2024b), identify broadly comparable competence areas despite differences in terminology. Strategic (methodological) competence coordinates the translation process and governs decision-making; thematic competence provides knowledge of legal systems, legal concepts and legal reasoning; linguistic and textual competence enables translators to understand source texts and produce idiomatic target texts; instrumental competence covers efficient technology use and research while professional and interpersonal competence governs ethical and organisational aspects of practice. Among these models, Prieto Ramos (2024b) already recognises that machine translation and post-editing cannot be confined to technological competence but depends primarily on strategic decision-making, thematic and linguistic competences.

Strategic competence is most directly affected because it governs decisions about whether AI should be used, which technologies are appropriate for a particular assignment, and what level of human intervention is required. These decisions depend on the risk profile of the task, client requirements and professional obligations. Strategic competence therefore integrates the highest level of AI literacy by directing the responsible use of AI throughout the translation process. Thematic together with linguistic/textual competences provide the expertise required to verify AI-generated output. They enable translators to assess legal accuracy, legal effects, terminological consistency, and naturalness of target texts, while identifying hallucinations, inaccuracies and automation bias. Instrumental competence encompasses the effective integration of AI into translation workflows. It also covers knowledge of AI systems and their limitations, secure workflow configuration, prompt engineering, AI-assisted terminology research, information retrieval, drafting and revision, as well as the ability to combine NMT, LLMs and other technologies appropriately within hybrid workflows. Professional and interpersonal competence includes compliance with legal and regulatory requirements governing AI use, transparent communication with clients, and the capacity to adapt to technological change through continuing professional development.

AI literacy should therefore be understood as a transversal framework that reshapes how existing legal translation competences are exercised in AI-assisted environments. It strengthens the capacity of competence models to address the AI-risk profile of legal translation while preserving the central role of professional judgement.

## 4. GenAI in legal translation classroom: Building trainees' AI literacy and digital resilience

The emergence of AI-assisted legal translation calls for corresponding changes in legal translator education by incorporating AI literacy as a learning outcome. This need has acquired an additional regulatory dimension under the EU AI Act, which requires deployers of AI systems to possess a sufficient level of AI literacy. As translation graduates entering professional practice will increasingly become such deployers, translator education has a responsibility to equip them with AI literacy alongside the core translation competences. This section examines how AI literacy can be developed in legal translation education. It begins by reporting how trainees use GenAI, discusses pedagogical principles and proposes classroom activities designed to develop AI literacy and digital resilience.

### 4.1 How students use GenAI?

To obtain a diagnostic picture of students' AI practices, I conducted a pilot survey based on open-ended questions in March 2026 among first-year MA students (n=22) enrolled in the Applied Linguistics (Translation and Translation Technologies) programme at the University of Warsaw. The findings indicate that students generally adopt a cautious and pragmatic approach to GenAI, which is broadly consistent with recent surveys of professional translators (Rivas Ginel and Moorkens (2025), Mao and Zheng (2026)).

Students actively combine general-purpose LLMs with specialised NMT to handle different stages of the translation workflow. For producing draft translations of legal or specialised texts, most respondents rely on NMT, in particular DeepL, frequently comparing outputs across platforms. By contrast, general-purpose LLMs like ChatGPT, Gemini, and Claude are used primarily for supporting task, including terminology research, searching for synonyms and cultural equivalents, improving style and fluency, proofreading, and conducting background research, especially on niche topics or when authoritative resources are scarce. A small minority of students reported not using LLMs to draft translations.

The survey further reveals that students have begun to develop prompting skills, although these remain at an introductory level. Prompts typically consist of straightforward zero-shot requests (e.g. *Translate this into Polish/English*) or few-shot prompts with limited contextual information. More advanced prompting strategies and iterative interactions with LLMs are uncommon. At the same time, respondents demonstrate some awareness of the limitations of LLMs. They report verifying AI-generated terminology and translation solutions against authoritative resources, including institutional websites, terminology databases (e.g. IATE), corpora, dictionaries and parallel texts.

Overall, these findings suggest that students are already incorporating GenAI into their workflows with some degree of critical awareness, yet mostly through ad hoc practices developed independently. This highlights the need for pedagogical interventions that develop AI literacy in a structured manner.

### 4.2 Developing AI literacy in legal translator education: Curriculum sequencing and pedagogical principles

It is by now a commonplace observation that integrating GenAI into translator education should extend beyond teaching students how to operate AI systems. As AI technologies evolve rapidly, students usually acquire basic operational skills independently. The primary objective of university education is therefore to develop the professional judgment required to use AI competently throughout the translation process.

A key pedagogical principle is curriculum sequencing. Students should first develop legal translation methodology and the core translation competences through human from-scratch translation before progressing to AI-assisted workflows. Premature reliance on AI-generated output may weaken the acquisition of these competences and encourage uncritical acceptance of automated solutions. This is consistent with research on machine translation pedagogy, which shows that post-editing produces better learning outcomes once learners have acquired initial translation experience (Setkowicz-Ryszka 2025). A curriculum should therefore progress from human translation supported by CAT tools towards AI-assisted workflows.

This progression also mirrors the AI literacy framework proposed in Section 3.2. The foundational, procedural and critical dimensions develop progressively as students acquire both legal translation expertise and experience with AI-assisted translation. Initially, AI should augment human translation decision-making by supporting source-text analysis, terminology research and the solution of micro-level translation problems. Once students have developed sufficient legal translation competence and verification routines, AI should be used for higher levels of automation, including drafting complete target texts under human oversight. Throughout this progression, strategic AI literacy must be present from the outset. Decisions concerning whether AI should be used, which system is appropriate and how AI-generated output should be verified precede interaction with AI and guide the entire translation process.

Prompting has become an important new skill in translator education (cf. Yamada (2025), Inglada (2025)). Research suggests that well-designed prompts can improve LLM performance in legal translation (e.g. Aldosari and Altuwairesh (2025), Hu (2026))). While prompting can be viewed simply as a technique to optimise output quality, from a pedagogical perspective it can serve as a bridge between translation theory and practice (Yamada 2025). Moving beyond simple zero-shot prompts, students learn to formulate context-sensitive prompts that incorporate elements of translation theory and/or methodology. Such interactions require students to articulate translation decisions and evaluate whether the AI output reflects them.

Another challenge concerns the changing role of AI in information-mining for translation tasks. Classroom observations indicate that students increasingly use LLMs as search engines and, hence, as sources of knowledge and expertise. This reflects a broader paradigm shift in how people access and evaluate information in the AI era (Reyes and Ross 2025, 2). In legal translation, however, students must learn to distinguish between AI-generated information and legally authoritative sources, understand the hierarchy of legal sources, and treat AI-generated explanations, terminology and references as provisional hypotheses requiring verification. While Inglada (2025, 52) argues that training should focus on assessing whether AI output is suitable (fit for purpose) rather than correct, legal translation requires a stricter standard: suitability depends on legal and factual correctness, which remains the responsibility of the translator.

The following section illustrates how these pedagogical principles can be implemented through classroom activities aligned with the four dimensions of AI literacy.

### 4.3 Classroom activities for developing AI literacy

The pedagogical principles outlined above can be translated into classroom activities integrated across the successive stages of the legal translation process (Table 2). The activities follow the progression proposed in the AI literacy framework. Foundational, procedural and critical AI literacy develop progressively as students move from source-text analysis to translation and (self-)revision, while the strategic dimension accompanies every stage by guiding decisions about AI use, technology selection and risk management. Thus, AI is introduced initially as a tool that augments human translation decision-making before being employed for increasingly automated translation tasks under human oversight.

Table 2. Classroom activities for developing AI literacy organised according to translation process stages.

| Stage | Classroom activity | Learning objective | AI literacy |
|---|---|---|---|
| Pre-translation | Task and brief analysis: workflow planning | Decide whether AI could be used; select appropriate technology; understand how AI systems operate; discuss advantages and limitations of AI use | Strategic, Foundational |
| | AI-or-human risk assessment across task types (e.g. legal acts, contracts, sworn translation) | Identify risks; develop awareness of constraints governing AI use in legal translation | Strategic, Foundational |
| | Client communication scenarios | Explain AI-related risks and justify professional decisions to clients | Strategic |
| | AI-assisted source-text contextualisation | Improve understanding of legal genre, communicative purpose and institutional context | Procedural |
| | Gisting and simplification of source texts | Support source-text comprehension before translating text fragments | Procedural |
| | AI-assisted terminology extraction, definition and equivalent generation followed by verification against authoritative legal sources | Support source-text comprehension; develop terminology research skills and verification routines; distinguish AI-generated information from authoritative sources; understand strengths and limitations of LLMs as research assistants | Procedural, Critical |
| Translation | Prompt engineering workshop (e.g. few-shot, chain-of-thought, persona prompting, context-rich and theory-informed prompting) | Develop effective interaction with AI and understand the effect of prompting on output quality; apply the knowledge of legal translation methodology to improve AI performance | Procedural, Critical |
| | AI assistance in micro-level tasks: source-text disambiguation, search for synonyms and equivalents, stylistic refinement | Use AI effectively within legal translation workflows; analyse, assess and implement AI-suggested solutions; establish verification routines | Procedural, Critical |
| | Controlled terminology experiments (with and without glossaries) | Demonstrate the impact of terminology control on AI performance | Procedural, Critical |

| | Comparative translation: human vs NMT vs LLM | Analyse differences in quality and adequacy; understanding differences between human and automated translation | Critical |
|---|---|---|---|
| | Comparison of outputs generated by different LLMs | Demonstrate output variability and train critical evaluation | Critical |
| | AI-assisted target-text generation | Generate draft translations under human oversight; analyse and edit AI outputs; identify hallucinations, mistranslations and other quality issues | Procedural, Critical |
| **Post-translation** | AI-assisted self-revision | Use AI as a quality assurance tool | Critical |
| | AI-assisted revision and post-editing | Develop revision and post-editing routines; detect quality issues | Critical |
| | Linguistic quality checking and stylistic revision L1/L2 | Improve translation quality | Critical |
| | AI-supported peer evaluation | Compare revision strategies and discuss translation decisions collaboratively | Critical, Strategic |
| | Reflective commentary on AI use | Justify AI-use decisions across the workflow; explain prompting and verification strategies; reflect on risk holistically; encourage reflective practice | Strategic |

During the pre-translation stage, students use LLMs primarily to support understanding of the source text. Activities include contextualisation, gisting (in particular, when working on fragments), extracting terminology and definitions, and conducting preliminary legal research. AI-generated information is verified against authoritative legal sources to establish systematic verification routines. Students also analyse translation briefs and assess whether AI is appropriate for particular assignments in light of risks and client requirements. At this stage, AI becomes an assistant for informed preparation.

During the translation stage, students progressively expand AI use from supporting individual translation decisions to drafting target texts. Initially, AI assists with micro-level tasks such as source-text disambiguation, terminology searches, equivalent selection or stylistic refinement. To develop the procedural and critical dimensions of AI literacy, students compare human translations with NMT and LLM output, analyse differences between competing LLMs, investigate the effects of terminology control and prompting strategies on translation quality, and identify quality issues in AI outputs. Prompt engineering exercises encourage more sophisticated interaction with AI through context-rich and theory-informed prompts that incorporate translation strategies and techniques, legal context, target readership or terminological constraints. Finally, as students develop critical evaluation skills, AI is introduced for automatic target-text drafting under human oversight.

During the post-translation stage, the emphasis shifts to verification and reflection. Students use AI selectively to support (self)-revision, improve linguistic quality and readability of draft target texts, particularly when translating into L2, or to post-edit machine translations. Reflective commentaries require students to take a step back and evaluate their workflow from a holistic, bird's-eye perspective, justifying technology choices, prompting strategies, verification procedures and overall risk assessment. These activities consolidate strategic AI literacy by linking AI use with the methodology and professional standards of legal translation.

### 4.4 From AI literacy to digital resilience

When developed systematically across the four dimensions, AI literacy becomes the basis of digital resilience. It enables future legal translators to adapt to technological change without relinquishing professional autonomy or becoming overdependent on automated systems. This resilience depends equally on strong legal translation competence. Knowledge of legal systems, legal language, terminology, genre conventions and professional standards remains indispensable for evaluating AI-generated output and determining whether it is acceptable.

Digital resilience also has a psychological dimension. Rapid technological change has generated digital anxiety, mainly in the form of concerns among translation students regarding employability and the future of the profession. Legal translator education should address these concerns by presenting AI as an augmentation of professional practice. Engagement with practising professionals, discussion of emerging roles in multilingual legal communication, and development of translation competence combined with AI literacy can strengthen students' confidence and adaptive capacity.

## 6. Conclusion

This chapter has argued that the growing use of GenAI does not alter the fundamental objectives of legal translation, but changes the conditions under which they are achieved. AI-assisted workflows introduce linguistic, technical, legal, ethical and cognitive risks. To address these challenges, the chapter has proposed an AI literacy framework for legal translation organised around four mutually reinforcing dimensions — foundational, procedural, critical and strategic. The framework adapts general models of AI literacy to the requirements of legal translation by placing professional judgement and risk management at its centre. The chapter further demonstrates how the AI literacy framework can be integrated into legal translator education through targeted pedagogical interventions.

The framework proposed here requires empirical validation. Future research should examine how AI literacy develops in legal translator education, evaluate the effectiveness of pedagogical interventions, and investigate methods for assessing AI literacy and digital resilience. Further work is also needed on prompting strategies for legal translation, human verification methodologies, and the evolving relationship between AI literacy and legal translation competence.